\documentclass[letterpaper, 10 pt, conference]{ieeeconf}  
\usepackage[bindingoffset=0.2in,%
            left=.667in,right=.667in,top=.792in,bottom=.597in,%
            footskip=.75in]{geometry}

\IEEEoverridecommandlockouts                              

\title{\LARGE \bf
Attentional-GCNN: Adaptive Pedestrian Trajectory Prediction towards Generic Autonomous Vehicle Use Cases
}

\author{Kunming Li$^{1}$, Stuart Eiffert$^{1}$, Mao Shan$^{1}$, Francisco Gomez-Donoso$^{2}$, Stewart Worrall$^{1}$ and Eduardo Nebot$^{1}$
\thanks{This  work  has  been  funded  by  the  Australian  Centre  for Field Robotics (ACFR), University of Michigan / Ford Motors Company Contract ``Next generation Vehicles”,  Transport for New South Wales (TfNSW) and iMOVE CRC and supported by the  Cooperative  Research  Centres  program, an Australian Government initiative.}
\thanks{$^{1}$Australian Centre for Field Robotics, University of Sydney, Sydney, NSW 2006, Australia.
        {\tt\footnotesize  \{k.li, s.eiffert,m.shan,s.worrall,e.nebot\}@acfr.usyd.edu.au}.}%
\thanks{$^{2}$University Institute for Computer Research, Universidad de Alicante, Spain.
        {\tt\footnotesize fgomez@ua.es}}%
}

\usepackage{gensymb}
\usepackage{times}
\usepackage{epsfig}
\usepackage{graphicx}
\usepackage{amsmath}
\usepackage{amsfonts}
\usepackage{amssymb}
\usepackage{bm}
\usepackage{comment}
\usepackage{longtable}
\usepackage{hhline}
\usepackage{ctable} 
\usepackage{multirow}
\usepackage{caption}
\usepackage{cite}

\usepackage{xcolor}

\begin{document}

\maketitle
\thispagestyle{empty}
\pagestyle{empty}

\begin{abstract}
Autonomous vehicle navigation in shared pedestrian environments requires the ability to predict future crowd motion both accurately and with minimal delay. Understanding the uncertainty of the prediction is also crucial. Most existing approaches however can only estimate uncertainty through repeated sampling of generative models. Additionally, most current predictive models are trained on datasets that assume complete observability of the crowd using an aerial view. These are generally not representative of real-world usage from a vehicle perspective, and can lead to the underestimation of uncertainty bounds when the on-board sensors are occluded.
Inspired by prior work in motion prediction using spatio-temporal graphs, we propose a novel Graph Convolutional Neural Network (GCNN)-based approach, Attentional-GCNN, which aggregates information of implicit interaction between pedestrians in a crowd by assigning attention weight in edges of the graph. 
Our model can be trained to either output a probabilistic distribution or faster deterministic prediction, demonstrating applicability to autonomous vehicle use cases where either speed or accuracy with uncertainty bounds are required.
 To further improve the training of predictive models, we propose an automatically labelled pedestrian dataset collected from an intelligent vehicle platform representative of real-world use. Through experiments on a number of datasets, we show our proposed method achieves an improvement over the state of art by $10\%$ Average Displacement Error (ADE) and $12\%$ Final Displacement Error (FDE) with fast inference speeds.

\end{abstract}


\section{INTRODUCTION}
Pedestrian motion prediction in crowds remains a significant challenge due to the complexity of modelling social interactions. This challenge is made increasingly difficult when implemented for the purposes of mobile robot or autonomous vehicle navigation. These applications require the consideration both of a predictive model's inference speed and accuracy, as well as an understanding of the limited ability of the vehicle to observe its surroundings.

\begin{figure}
    \centering
	\includegraphics[width=8.5 cm,height=5.5cm]{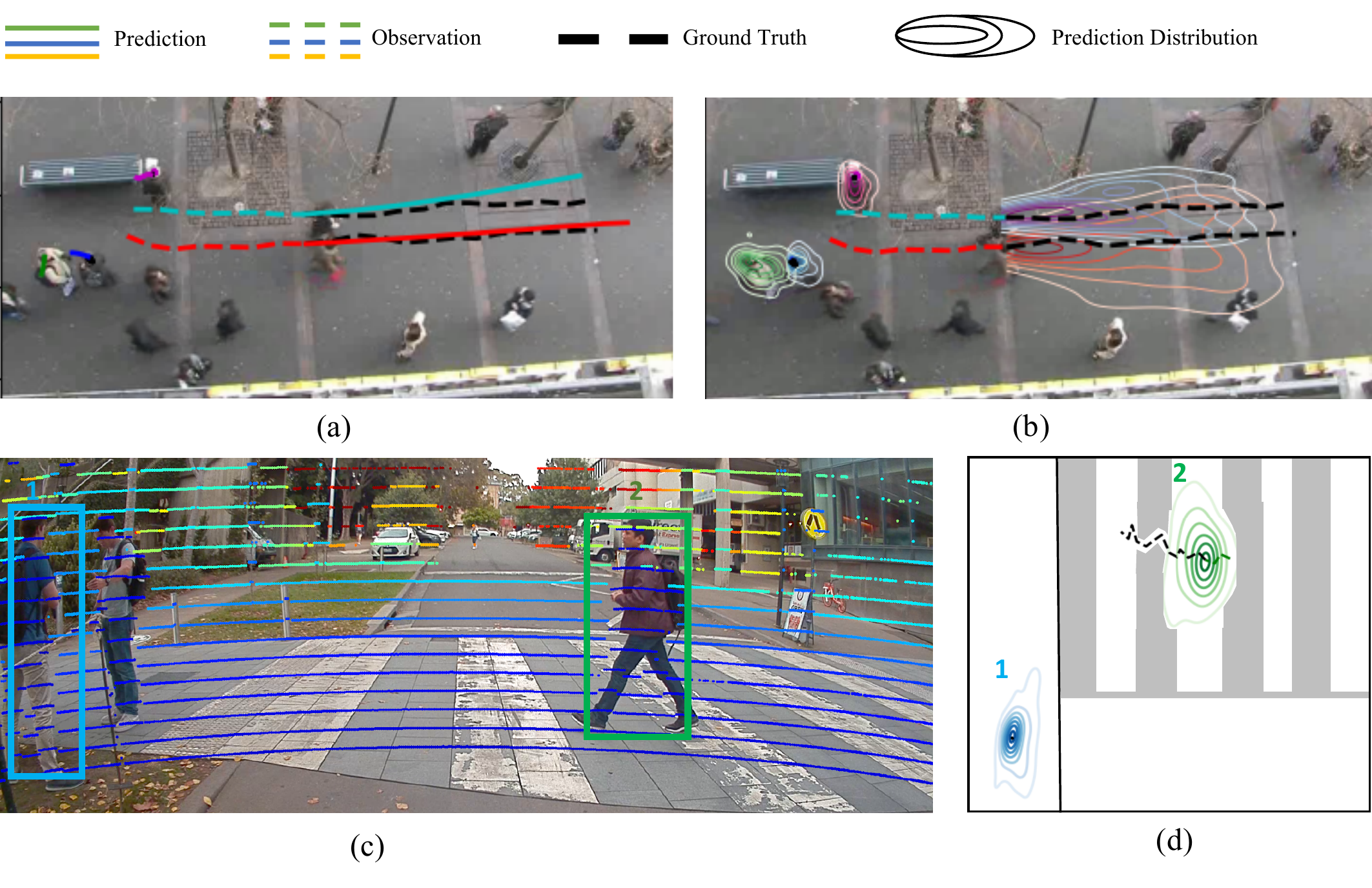}
	\caption{\textit{Results of our proposed method on top-down (top) or on-vehicle (bottom) data. (a) and (b) are results of our proposed deterministic model and probabilistic model respectively, with distributions shown as contour maps generated from 20 samples. (c) and (d) show the probabilistic prediction result on our proposed USyd Pedestrian dataset.}}
	\label{front_image}
\end{figure}

Minor delays between when a vehicle observes a crowd and when the vehicle acts can result in significant differences in crowd states, resulting in the planned action no longer being valid. 
As such, consideration of not just model prediction accuracy, but also model inference speed is crucial to allow autonomous vehicles to safely and effectively operate in shared environments alongside pedestrians. 

Most recent crowd motion prediction models can be categorised as either deterministic \cite{alahi2016social, zhang2019sr, Vemula2018} or probabilistic with regards to their outputs. Whilst some probabilistic approaches directly output a distribution \cite{eiffert2020probabilistic}, many of the highest performing methods require repeated sampling in order to approximate the true distribution of crowd trajectories \cite{gupta2018social, ivanovic2019, sophie2018, liang2019peeking, zhang2019stochastic, kosaraju2019social}. The prediction uncertainty provided by the full distribution can be especially important in complex environments such as those with dense crowds, where pedestrians tend to change their walking direction or speed abruptly in order to avoid collisions. However, repeated sampling can lead to longer inference times than deterministic methods, creating a clear trade-off between accuracy and speed in safety-critical applications.

Additionally, perception in real-world robotic applications are generally limited to available on-board sensors, such as 2D cameras and lidar. Most existing predictive models such as  \cite{alahi2016social, zhang2019sr, Vemula2018,gupta2018social,sophie2018,zhang2019stochastic,kosaraju2019social} are trained on datasets which are collected from top-down views and assume that the vehicle can fully observe a crowd. 
These models have not been trained on datasets representative of a vehicles perspective, in which pedestrians are often occluded from view. As such, they may fail to predict the possible reaction of the detected pedestrian to unobserved pedestrians.


In this work, we propose a novel pedestrian trajectory prediction approach, namely Attentional-GCNN, which takes into consideration the inference speed, accuracy, and real-world limitations of a predictive model during autonomous vehicle navigation. The proposed GCNN-based pedestrian prediction model can be trained to either output the full distribution of future trajectories, or a deterministic output, dependent on the desired use case. 
We introduce a novel Near Pedestrian Attention (NPA) function that improves prediction accuracy during crowd interactions by embedding information of mutual influence between pedestrians. Our approach achieves both improved inference speed and smaller model size than previous Recurrent Neural Network (RNN) based pedestrian prediction methods. We further propose an automatically labelled pedestrian dataset collected from an intelligent vehicle platform representative of a real-world vehicle application, in order to train, verify and evaluate predictive models implemented for autonomous vehicle navigation.


\section{Related Work}
\textbf{Deterministic and Probabilistic Models}\newline
Pedestrian motion prediction plays an essential role in a variety of autonomous driving tasks.
Time-critical tasks such as collision avoidance often make use of fast deterministic models of pedestrian motion \cite{Rudenko2020,Magdici2016}. Probabilistic models, which either directly output a distribution per pedestrian \cite{eiffert2020probabilistic, ivanovic2019,Salzmann2020, mohamed2020social}, or sample from a learnt distribution \cite{gupta2018social,kosaraju2019social}, are instead used in tasks requiring an understanding of prediction uncertainty. This includes interactive tasks such as sampling-based path planning in dense crowds \cite{Eiffert2020a} or validating the safety of existing plans \cite{Salzmann2020 , Fisac2018}. 

Current state of the art approaches to pedestrian motion prediction tend to make use of RNNs for both probabilistic and deterministic applications, modelling social interaction via pooling layers across RNN hidden states \cite{alahi2016social, zhang2019sr, eiffert2020probabilistic}. Graph Attention Networks \cite{velickovic2018graph} have also shown how attention can be encoded in these pooling layers, achieving improved performance in dense interactions \cite{kosaraju2019social, eiffert2020probabilistic}. Similarly, Spatio-temporal graphs have been used to encode social interactions in model inputs \cite{Vemula2018,Eiffert2019,ma2019trafficpredict}. Generative Adversarial Networks (GANs) have been shown to improve pedestrian trajectory prediction accuracy, however, these approaches are limited by an inability to directly output a prediction as a distribution, requiring repeated stochastic sampling of the model to reveal the approximate distribution \cite{gupta2018social,kosaraju2019social}. Recently, \cite{eiffert2020probabilistic} has shown how this can be extended to directly output the distribution by utilising Gaussian mixture models as a generator output and modal-path clustering before making a discriminator comparison.

\textbf{CNN based trajectory prediction}\newline
Recent work has shown that CNN based approaches can achieve comparable performance to RNNs in sequence prediction but with much reduced computational resources \cite{mohamed2020social, wang2020joint, bai2018empirical, vaswani2017attention}. Social interactions in CNN based approaches have been modelled with similar methods as RNNs by using spatio-temporal graph inputs and Graph Neural Networks (GNN) \cite{kipf2016semi} to allow CNNs to operate directly over a graph domain as Spatio-Temporal Graph Convolution Neural Networks (ST-GCNN) \cite{mohamed2020social}. We extend this approach by constructing the graph to encode attention between pedestrians through our NPA function, detailed in Section 3.

\textbf{Usage Representative Training Datasets}\newline
Most existing datasets used to train pedestrian motion prediction models are collected from top-down views that provide clear and continuous pedestrian trajectories with complete information of a scene\cite{lerner2007crowds,robicquet2016learning, yang2019top,benfold2011stable,zhou2012understanding}. However, in real-world vehicle applications, observations are generally limited to on-board cameras and lidar resulting in frequent occlusions and incomplete crowd observations. Predictive models trained on fully observable datasets will be less accurate when a detected pedestrian is reacting to other unobserved pedestrians. By instead training using only the observable pedestrians as input, the models will better reflect real-world use. The ground truth motion of each agent will still incorporate the responses to other unobserved pedestrians. As a result, these models will assign greater, more realistic uncertainty to predictions, and potentially better learn to predict trajectories in partially observed crowds. 

Several on-vehicle datasets exists \cite{sun2020scalability,geiger2012we, houston2020one} which focus on pedestrian-vehicle or vehicle-vehicle interactions. Our proposed on-vehicle pedestrian dataset instead focuses on social interactions between pedestrians in the presence of a vehicle. A comparison of datasets is shown in \textit{Table \ref{table_dataset}}.

\begin{table}[] 
\setlength\tabcolsep{4.5 pt} 
\begin{tabular}{c|c|c|c}
\hline
Dataset Name            & \multicolumn{1}{l|}{Pedestrians Only} & View     & Sensor        \\ \hline \hline
ETH                    & Yes                                     & Top-down & camera        \\ \hline
UCY                     & Yes                                     & Top-down & camera        \\ \hline
DUT                    & No                                    & Top-down & camera        \\ \hline
Standford               & No                                    & Top-down & camera        \\ \hline
Town center             & No                                    & Top-down & camera        \\ \hline
Waymo Open Dataset      & No                                    & On-vehicle    & camera, lidar \\ \hline
Lyft Prediction Dataset & No                                    & On-vehicle    & camera, lidar \\ \hline
USyd Pedestrian Dataset & Yes                                     & On-vehicle    & camera, lidar \\ \hline
\end{tabular}
\caption{\textit{Comparison of existing pedestrian datasets for autonomous driving systems. We compare whether data was captured from on-vehicle, representative of real-world usage, or from a top-down view.}\strut}
\label{table_dataset}
\end{table}

\section{Method}
\subsection{Problem Definition}
Given observed pedestrian trajectories $\textbf{X}$ from all time steps in period $ t \leq T_{obs}$, where $\textbf{X}^t= [\textbf{X}_1^t,\textbf{X}_2^t...,\textbf{X}_N^t]$ for $N$ pedestrians in a scene, our goal is to predict future trajectories $\hat{\textbf{Y}}^t= [\hat{\textbf{Y}}_1^t,\hat{\textbf{Y}}_2^t...,\hat{\textbf{Y}}_N^t]$ over the prediction time period $ T_{obs} < t \leq T_{pred}$.

The  input  position  of  the $i$th pedestrian  at  time $t$ is  denoted  as $\textbf{X}^t_ i=  (x^t_i,y^t_i)$, and in ground truth future trajectories $\textbf{Y}$ as $\textbf{Y}^t_i=  (x^ti,y^t_i)$. During training, $\textbf{Y}$ is compared to the predictive model's output $\hat{\textbf{Y}}$. This output takes the form of a bi-variate Gaussian distribution $\hat{\textbf{Y}}_{i}^t \sim  \mathcal{N}( \hat{\mu}^{t}_{i},\,\hat{\sigma}^{t}_{i},\hat{\rho}^{t}_{i})$ over the positional $x$ and $y$ dimensions in the probabilistic model version, or $\hat{\textbf{Y}}^t_i = (x^ti,y^t_i)$ in the deterministic version.

\begin{figure*}[tbp] 
\centering
\includegraphics[width=17.5cm,height=4.375cm]{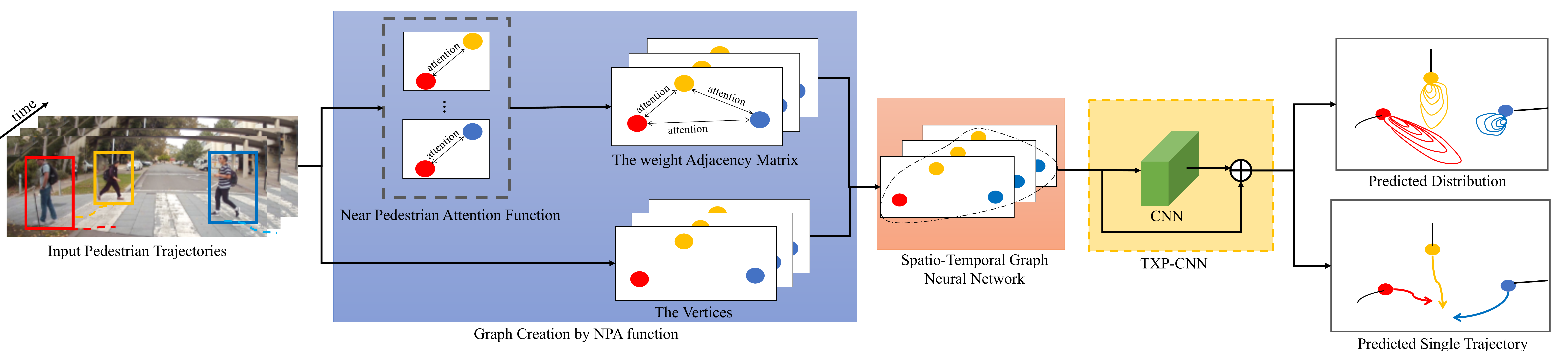}
\caption{\textit{Overview of our proposed method. Observed pedestrian trajectories are constructed as the vertices of a graph, with the weight adjacency matrix computed by Near Pedestrian Attention function, which are passed to Spatial Temporal Network. TXP-CNN uses residual blocks to extrapolate pedestrians' future trajectories. Our proposed method is adapted to produce either a distribution of trajectories or the single most likely trajectory through different loss functions.}}
\label{fig_archi}
\end{figure*}

\subsection{Model Description}
Our proposed model consists of three parts: (1) a Near Pedestrian Attention Function (NPA); (2) the Spatio-Temporal Graph Convolution Neural Network (ST-GCNN); (3) and the Time-Extrapolator Convolution Neural Network (TXP-CNN), as shown in Fig. \ref{fig_archi}.


\subsubsection{\textbf{Graph Construction with Near Pedestrian Attention Function}} \label{sec_npa}

For an observed sequence, we construct a graph representation of the scene using our NPA function. At each timestep, we define an undirected graph $ \textbf{G}_{t}  = ( \textbf{V}_{t}, \textbf{A}_{t})$ where each node $v_{t}^{i} \in \textbf{V}_{t} $ represents the displacement of agent $i$ from the prior timestep, for $N$ pedestrians. Edges of the graph are represented using a weighted adjacency matrix $ \textbf{A}_{t})$, where each edge $a^{i,j}_{t} \in \textbf{A}_{t})$ represents the attention between neighbours $i$ and $j$. 
In this work, we assume that a pedestrian's motion is more easily influenced by nearby neighbours, and use our NPA function to compute attention. Closer neighbours are assigned higher attention weight by softmax as \textit{Eq.\ref{eq_dis}} and \textit{Eq.\ref{eq_softmax}} show.

\begin{figure*}[tbp] 
\centering
\includegraphics[width=17.5cm,height=6.0cm]{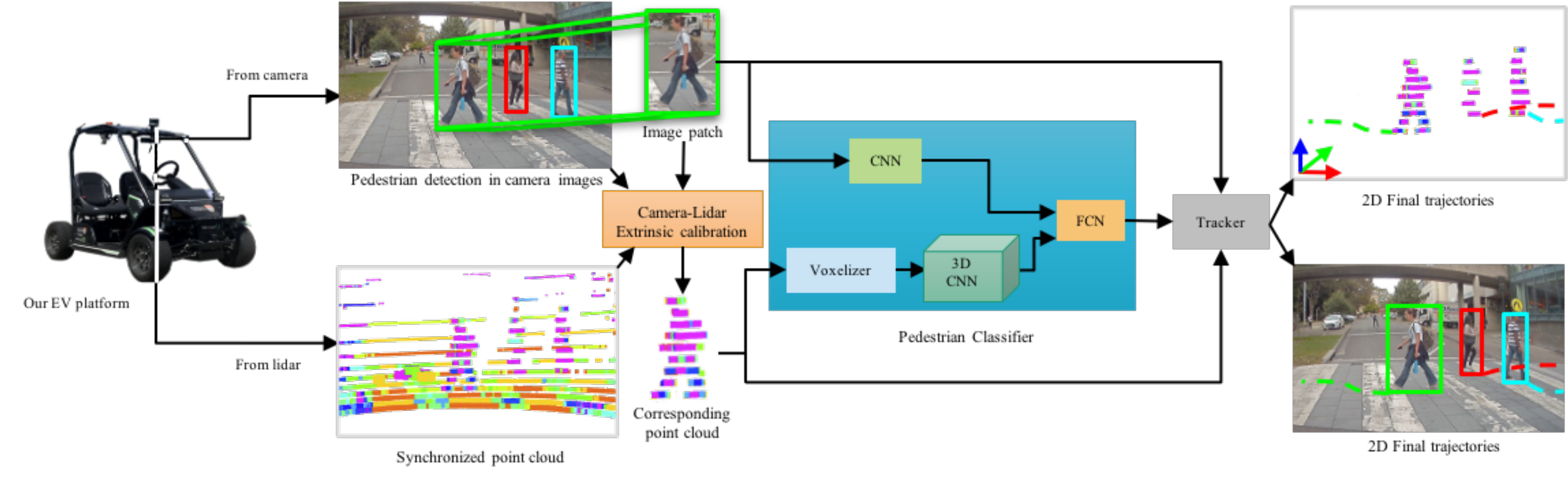}
\caption{\textit{Overview  of the automatic labelling pipeline used to create Usyd Pedestrian dataset. Lidar points and images are captured by sensors mounted on the EV platform. 2D detections are associated pointcloud clusters and classified as either true pedestrians or false positives using the Pedestrian Classifier. True pedestrians are then tracked and transformed into world coordinates, as described in Section \ref{section:labelling}.}}
\label{fig_perception}
\end{figure*}

\begin{align}
d^{t}_{i,j} = \| (x^{t}_{i}-x^{t}_{j},y^{t}_{i}-y^{t}_{j}) \| \label{eq_dis} \\ 
a^{t}_{i,j} = \frac{exp(d_{i,j}^{t})}{\sum\limits_{k \in N \setminus \{ i\}} exp(d_{i,k}^{t})} \label{eq_softmax}
\end{align}
The publicly available implementation of Social-STGCNN \cite{mohamed2020social} details usage of relative velocity between neighbours as input during graph adjacency matrix creation. In this work, we demonstrate that by instead using the relative distances between neighbours, scaled using a softmax operation to represent attention, we can better represent the influence of nearby neighbours on a pedestrian during interactions. 

We use the entire spatio-temporal graph $ \textbf{G}=	\{\textbf{G}_1,..,\textbf{G}_{T_{obs}}\}$ across the observed time period $T_{obs}$ as input to the ST-GCNN.

\subsubsection{\textbf{Spatio-Temporal Graph Convolution Neural Network}}
We apply the graph spatial convolution operation, first introduced in \cite{mohamed2020social}, to $\textbf{G}$. This generalizes the operation of convolution from an array to graph representation, using a kernel to aggregate features across a local neighbourhood of the graph in both spatial and temporal dimensions in a similar manner as grid based CNNs. Following prior work \cite{kipf2016semi} \cite{mohamed2020social}, we also normalize the adjacency matrix $\textbf{A}$ for each timestep $t$ to allow proper feature extraction:
\begin{align}
\hat{\textbf{A}}_{t} = \tilde{\textbf{D}}_{t}^{ -\frac{1}{2}} (\textbf{A}_{t}+\textbf{I})\tilde{\textbf{D}}_{t}^{ -\frac{1}{2}} \label{eq_norm}
\end{align}
where $\tilde{\textbf{D}}_{t}$ is denoted as the diagonal node degree of $(\textbf{A}_{t}+\textbf{I})$ at time step $t$ and $\textbf{I}$ is denoted as the identity matrix. $\hat{ \textbf{A}}_{t}$ presents the normalized weight adjacency matrix at time $t$.

The input to each layer $l$ of the ST-GCNN is described below, where $\textbf{V}$ and $\hat{ \textbf{A}}$ denote the concatenations of $\textbf{V}_t$ and $\hat{ \textbf{A}}_t$ along the time dimension respectively, for all $t \in T_{obs}$. The input to the first layer $H^{0}$ is $\textbf{V}$, whilst $\hat{\textbf{A}}$ is used for all layers:  
\begin{align}
    H^{(l+1)} = \sigma(\hat{\textbf{A}} H^{(l)}W^{(l)}) \label{eq_stgcnn}
\end{align}

$W^{l}$ is a layer-specific trainable weight matrix and $\sigma$ is the activation function.

\subsubsection{\textbf{Time-Extrapolator Convolution Neural Network}}
 TXP-CNN, described in \cite{mohamed2020social}, acts to decode the final embedded output $\textbf{H}$ from ST-GCNN along the temporal dimension, allowing prediction of future pedestrian trajectories for a given time period $T_{pred}$. This module was inspired by Temporal Convolution Neural Networks (TCNs)\cite{bai2018empirical} and makes use of $K$ residual layers \cite{he2016deep}. The final output $\textbf{U}^K$ becomes predicted output $\hat{\textbf{Y}}$, as expressed below:

 \begin{align}
     \textbf{U}^{0} = \sigma_{txp} (\textbf{H} W^{0}_{txp}) \\
     \textbf{U}^{(k+1)} = \sigma_{txp}((\textbf{U}^{k}W^{k}_{txp}) + \textbf{U}^{k}) \quad (k>=0)
 \end{align}

where $\sigma_{txp}$ and $W_{txp}^k$ represent the activation function and layer-specific trainable weight matrices respectively.

\subsubsection{\textbf{Loss Functions}}
The probabilistic and deterministic versions of our model make use of different loss functions during training. $L_{lh}$ is used to train a probabilistic output, minimizing the negative log-likelihood of the ground truth $ \textbf{Y}$ given the output distribution of the model $\hat{\textbf{Y}} $, which is defined as:
\begin{align}
    L_{lh} =  -\sum_{t=T_{obs}+1}^{T_{pred}}\sum_{i}^{N} \log (P({Y_{i}^{t}}|\hat{Y_{i}^{t}}))
\end{align}

We introduce $L_{cde}$ to train a deterministic output, minimizing average displacement error and final displacement error of the ground truth and the predicted deterministic trajectory as shown in \textit{Eq. \ref{eq_cde}}, where $\alpha$ is a hyperparameter between $[0,1]$.  
\begin{align}
\begin{split}
    L_{cde} = \alpha \sum_{t=T_{obs}+1}^{T_{pred}}\sum_{i}^{N} \|  {Y_{i}^{t}} - \hat{Y_{i}^{t}} \|  + \\  
    \quad (1-\alpha)\sum_{i}^{N} \|  {Y_{i}^{T_{pred}}} - \hat{Y}_{i}^{T_{pred}} \|     \label{eq_cde}
\end{split}
\end{align}

\subsection{Implementation}
The model consists of 1 layer of ST-GCNN and 5 layers of TXP-CNN, where both temporal convolutions and graph convolutions use a kernel size of 3. The deterministic model and the probabilistic model both use PReLu activations \cite{he2015delving} and are trained for 150 epochs with the batch size of 128. We use Adam optimiser \cite{kingma2014adam} with learning rate at 0.0015 for training the deterministic model, however we use the SGD optimiser with learning rate at 0.01 in the training of the probabilistic model as this was seen to achieve better training results.

\section{USyd Campus Pedestrian dataset} \label{usyddataset_method}
\begin{figure}
    \centering
	\includegraphics[width=8.6cm,height=6.14cm]{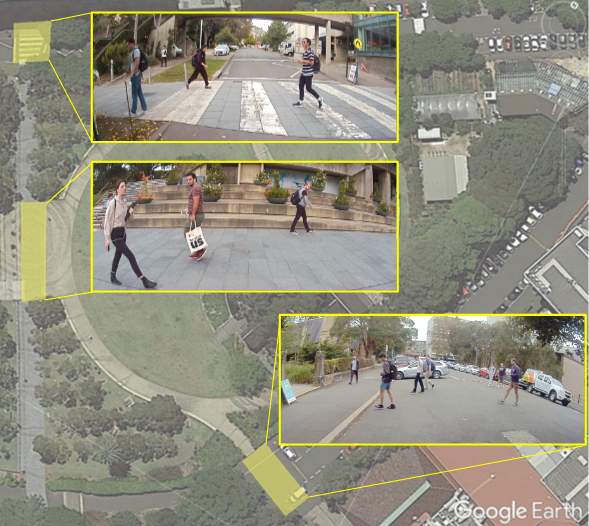}
	\caption{\textit{Google Earth map showing the locations used to collect the Usyd Pedestrian dataset.}}
	\label{data_layout}
\end{figure}

The USyd Campus Pedestrain dataset was collected at three different locations during peak times at the University of Sydney, as shown in Fig. \ref{data_layout} and contains more than 37 minutes in total. 

\subsection{Dataset Platform}
Data was captured from an Electric Vehicle (EV) platform similar to \cite{zhou2020developing}, as depicted in Fig \ref{fig_perception}. Sensing consists of three front NVIDIA 2Mega SF3322 automotive GMSL cameras and a 3D Velodyne Puck VLP-16 lidar. Lidar was captured at 10 Hz and RGB at 30Hz with a resolution of resolution of $1,928 \times 1,208 $. In testing, we downsample the raw camera data to 10 Hz to be consistent with the lidar sample rate. 2D cameras were intrinsically calibrated based on fish-eye camera model \cite{kannala2004generic} before extrinsic calibration with lidar, performed automatically using 3D point and plane correspondences, proposed by Verma et al. \cite{verma2019automatic}.

\subsection{Automatic Labelling}
\label{section:labelling}
The dataset was labelled using an automatic perception pipeline shown in Fig \ref{fig_perception} based on the work proposed by \cite{gomez2020using}. This pipeline includes 2D object detection using YOLOv3 \cite{2018yolov3}, a 3D lidar classification model based on \cite{garcia2016pointnet} and \cite{Gomez_Donoso_2020_par3dnet}, and a Deepsort visual tracker \cite{wojke2017simple}.  
Each potential pedestrian image patch found by YOLOv3 is projected into the 3D pointcloud, associating it with a cluster of points that depicts the subject in the lidar pointcloud. This cluster is then voxelised and forwarded to a 3D CNN. The image patch is also passed through another CNN. Finally, the output of both convolutional pipelines are concatenated and passed to a fully connected layer that classifies the sample as either a true pedestrian or a false positive. We relied on this pipeline instead of a 2D-based one because the urban environment contains a range of objects that look like pedestrians but they are not pedestrians. For instance, we could find billboards, posters and digital displays depicting people. A 2D-based approach would eventually detect these objects as pedestrians and, thus, extracting a false positive. 
The original image patch is then encoded by secondary 2D CNN to extract features used by Deepsort, associating it in the 2D frame. For true pedestrians, the original image patch is then associated with prior tracks in the 2D frame using Deepsort  \cite{wojke2017simple}, before the 3D coordinates of the pointcloud cluster are used to transform the track into the world frame. Linear interpolation is applied to allow trajectories to be continuous across short occlusions.

\section{Experiments}
We conduct several experiments to validate our proposed probabilistic and deterministic models, comparing performance to a number of baselines using 3 real-world datasets. These include two with a top-down view; ETH \cite{pellegrini2009you} and UCY \cite{lerner2007crowds} datasets, and our proposed USyd Pedestrian Dataset collected from an intelligent vehicle reflecting a real-world vehicle sensor implementation. 
Additionally, we compare both parameter size and inference times for tested methods. 

\subsection{Datasets}
\textbf{ETH} \cite{pellegrini2009you} contains two scenes, ETH-Univ is collected from a University, and ETH-Hotel is collected in urban street. The dataset is converted to world coordinates with an observation frequency of 2.5 Hz.

\textbf{UCY} \cite{lerner2007crowds} includes UCY-Zara01, UCY-Zara01 and UCY-Hotel collected from an urban scene. The trajectories in the datasets are sampled every 0.4 seconds.

\textbf{USyd Pedestrian Dataset} is collected by a vehicle operating in a university campus as described in \textit{Sec. \ref{usyddataset_method}} sampled at 10 Hz.  

\subsection{Evaluation Metrics and Baselines}
\subsubsection{Metrics} Similar to prior work \cite{eiffert2020probabilistic,zhang2019sr}, we use Average Displacement Error (ADE) and Final Displacement Error(FDE) to evaluate the deterministic models. In evaluation of probabilistic models, we use Best-of-N (BoN) in the same manner as prior work \cite{gupta2018social} \cite{liang2019peeking} \cite{mohamed2020social}.

The metrics used are as follows:
\begin{itemize}
  \item Average Displacement Error (ADE): Average Euclidean distance between ground truth and prediction trajectories for all predicted time steps.
  \item Final Displacement Error (FDE): Euclidean distance between ground truth and prediction trajectories at the final predicted time step.
  \item  Best-of-N (BoN): The best sample with minimum ADE and FDE amongst N sampled trajectories from the output distribution.
\end{itemize}

\medskip
\subsubsection{Baseline} We compare our deterministic and probabilistic model versions separately to the following baselines. We additionally compare our probabilistic model when limited to a single `most-likely' prediction against all deterministic methods:  

\textbf{Deterministic Models:}
\begin{itemize}
    \item Lin: A linear regression over each dimension. 
    \item CVM \cite{Scholler2019}: Kalman filter based model.
    \item SR-LSTM \cite{zhang2019sr}: LSTM based deterministic using a State Refinement module. 
    \item PC-GAN \cite{eiffert2020probabilistic} : LSTM encoder-decoder trained adversarially with attention pooling.  
\end{itemize}

\textbf{Probalistic Models:}
\begin{itemize}
    \item SocialGAN \cite{gupta2018social}: LSTM encoder-decoder with a social pooling layer, trained adversarially.
    \item Sophie \cite{sophie2018}: LSTM network with a proposed a physical and social attention module, trained adversarially
    \item STSGN \cite{zhang2019stochastic}: LSTM with graph attention network for stochastic trajectory prediction.
    \item Social-BiGAT \cite{kosaraju2019social}: LSTM network with a Graph Attention Network (GAT) to encode the interaction information, trained adversarially.
    \item Social-STGCNN \cite{mohamed2020social}: Spatio-temporal graph based neural network.
\end{itemize}

\subsection{Methodology}
For both the ETH and UCY datasets we observe the last 8 timesteps (3.2 seconds) per trajectory and predict for the next 12 timesteps (4.8 seconds). For the Usyd Pedestrian dataset, as tracks tend to be of shorter length due to occlusion from the vehicles viewpoint, we only observe 4 timesteps (0.4s) per trajectory and predict for the next timesteps (1.2 seconds). We split all datasets into training, validation and testing sets, in ratios ${3}/{5}$, ${1}/{5}$ and ${1}/{5}$ respectively.

We additionally compare the parameter size and inference time of each of the predictive models, an important consideration for real-world applications of autonomous vehicles.

\begin{table*}[]
\centering
\begin{tabular}{c|cccccc}
\hline
\multirow{2}{*}{Probalistics Models}   & \multicolumn{6}{c}{\textbf{(1) ADE/ FDE (m), Best of 20 Samples}}                                                                                \\ \cline{2-7} 
                                       & ETH-Univ           & ETH-Hotel          & UCY-Univ           & UCY-ZARA1          & \multicolumn{1}{c|}{UCY-ZARA2}          & Average            \\ \hline
Social GAN  \cite{gupta2018social}     & 0.87/1.62          & 0.67/1.37          & 0.76/1.52          & 0.35/0.68          & \multicolumn{1}{c|}{0.42/0.84}          & 0.61/1.21          \\
SoPhie   \cite{sophie2018}             & 0.70/1.43          & 0.76/1.67          & 0.54/1.24          & \textbf{0.30}/0.63 & \multicolumn{1}{c|}{0.38/0.78}          & 0.54/1.15          \\
Next    \cite{liang2019peeking}        & 0.73/7.65          & \textbf{0.30}/0.59 & 0.60/1.27          & 0.38/0.81          & \multicolumn{1}{c|}{0.31/0.68}          & 0.46/1.00          \\
STSGN   \cite{zhang2019stochastic}     & 0.75/1.63          & 0.63/1.01          & 0.48/1.08          & \textbf{0.30}/0.65 & \multicolumn{1}{c|}{\textbf{0.26}/0.57} & 0.48/0.99          \\
Social-BiGAT \cite{kosaraju2019social} & 0.69/1.29          & 0.49/1.01          & 0.55/1.32          & 030/0.62           & \multicolumn{1}{c|}{0.36/0.75}          & 0.48/1.00          \\
Social-STGCNN \cite{mohamed2020social} & \textbf{0.64/1.11} & 0.49/0.85          & 0.44/0.79          & 0.34/\textbf{0.53} & \multicolumn{1}{c|}{0.30/0.48}          & 0.44/0.75          \\ \hline
Ours (probalistic model)               & 0.68/1.22          & 0.31/\textbf{0.41} & \textbf{0.39/0.69} & 0.34/0.55          & \multicolumn{1}{c|}{0.28/\textbf{0.44}} & \textbf{0.40/0.66} \\ \hline \hline
\multirow{2}{*}{Detrministic Models}   & \multicolumn{6}{c}{\textbf{(2) ADE/ FDE (m)}}                                                                                                    \\ \cline{2-7} 
                                       & ETH-Univ           & ETH-Hotel          & UCY-Univ           & UCY-ZARA1          & \multicolumn{1}{c|}{UCY-ZARA2}          & Average            \\ \hline
Linear                                 & 0.79/1.57          & 0.39/0.72          & 0.82/1.59          & 0.62/1.21          & \multicolumn{1}{c|}{0.77/1.48}          & 0.68/1.31          \\
CVM  \cite{Scholler2019}               & 0.70/1.34          & \textbf{0.33}/0.62 & 0.56/1.20          & 0.46/0.99          & \multicolumn{1}{c|}{0.35/0.75} & 0.48/0.98 \\
PCGAN \cite{Eiffert2019}               & 0.65/\textbf{1.25 }         & 0.64/1.40          & 0.57/1.24          & \textbf{0.40}/0.89          & \multicolumn{1}{c|}{0.34/0.77} & 0.52/1.11          \\
SR-LSTM \cite{zhang2019sr}             & \textbf{0.63/1.25} & 0.37/0.74 & \textbf{0.51/1.10} & 0.41/0.90          & \multicolumn{1}{c|}{\textbf{0.32}/0.70} & \textbf{0.45/0.94} \\
Social STGCNN (most-likely) \cite{mohamed2020social}           & 1.02/1.95          & 0.66/1.32          & 0.69/1.36          & 0.59/1.18          & \multicolumn{1}{c|}{0.44/0.90}          & 0.68/1.34          \\ \hline
Ours (probabilistic - most-likely)    & 0.98/1.88          & 0.55/1.22          & 0.61/1.21          & 0.54/1.09          & \multicolumn{1}{c|}{0.48/0.94}          & 0.63/1.26          \\
Ours (deterministic model)             & 1.0/1.8            & 0.38/\textbf{0.50} & 0.6/1.14  & 0.45/\textbf{0.81} & \multicolumn{1}{c|}{0.36/\textbf{0.66}} & 0.55/0.98 \\ \hline
\end{tabular}
\caption{\textit{Comparison of tested probabilistic and deterministic models by ADE/FDE. Probabilistic models (Top) are evaluated using Best-of-N=20 samples. Our probabilistic method outperforms all other methods on average in both ADE and FDE by over 10 $\%$. For deterministic results (Bottom), we see that our approach adapted for deterministic output is able to outperform the most-likely output of our probabilistic version, achieving comparable results to state of the art SR-LSTM \cite{zhang2019sr} and outperforming all methods for FDE in 3 out of 5 datasets. Models use 8 frames as input and predict for 12.}}
\label{table_pro_deter}
\end{table*}

\section{Result and Discussion}
\subsection{Quantitative Evaluation}
\textbf{Probabilistic Predictions:} \textit{Table \ref{table_pro_deter}} displays the comparison for all probabilistic methods on the ETH \cite{pellegrini2009you} and UCY \cite{lerner2007crowds} datasets. On average, our probabilistic model obtains better results than all other compared probabilistic models with $10\%$ improvement on ADE and $12\%$ improvement on FDE compared to the next best model Social-STGCN \cite{mohamed2020social}. This result demonstrates that our proposed NPA module, which embeds relative distance based attention between pedestrians into the graph, improves performance greatly. Whilst our probabilistic model only achieves the best accuracy on 4 out of 10 individual metrics, it is clear from the average score that our approach is better able to generalise between datasets than other compared methods. Additionally, we see that both our proposed method and Social-STGCNN outperform the compared LSTM methods \cite{liang2019peeking, gupta2018social, kosaraju2019social}, demonstrating that GNN and CNN based methods can achieve significantly better results than LSTM based methods in pedestrian motion prediction tasks. It is clear from these results that the spatial-temporal graph neural network is able to successfully extract the features relevant to implicit interaction between pedestrians. 

\textit{Table. \ref{table_usyd}} shows results from the Usyd pedestrian dataset.
Our proposed probabilistic model and Social-STGCNN significantly outperform all other methods, including the probabilistic SocialGAN method, although Social-STGCNN achieves slightly better results in ADE and FDE. Due to the limited observation period of 4 timesteps, social interactions are not as prominent, and so our proposed NPA function was likely not able to better represent relationships. Additionally, significant noise exists compared to top-down datasets, as seen in the bottom right of Fig. \ref{front_image}. This is typical of real-world data collected from a vehicle, resulting from occlusions, sensor noise, and lidar firing cycles mismatching with 2D input. The noise is also a likely explanation for the minimal difference seen between ADE and FDE for our method and Social-STGCNN. The ground truth path contains significant noise however the predicted path directly leads to the final ground truth position, resulting in lower final but higher average error.

\begin{table}[]
\centering
\begin{tabular}{c|c}
\hline
              & USyd Pedestrian Dataset \\ \hline \hline
Linear        & 0.57/0.93               \\ \hline
CVM \cite{Scholler2019}          & 1.1/1.76                \\ \hline
Social GAN \cite{gupta2018social}    & 1.24/2.32$^*$               \\ \hline
Social STGCNN \cite{mohamed2020social} & 0.30/0.29$^*$               \\ \hline
Ours-Probabilistic         & 0.32/0.31$^*$               \\ \hline
Ours-Deterministic          & 0.46/0.84               \\ \hline
\end{tabular}
\caption{\textit{Comparison of ADE/FDE on USyd Pedestrian Dataset. All models take 4 frames as input and predict the next 12 frames. $^*$ indicates 20 samples used for evaluation.}}
\label{table_usyd}
\end{table}

\begin{table}[]
\centering
\begin{tabular}{c|c|c}
\hline
Method  & Number of Parameters & Inference Time (ms) \\ \hline \hline
S-GAN  \cite{gupta2018social}  & 46.6K                      &  $91.6^*$           \\ \hline
S-P-GAN \cite{gupta2018social}   & 360.3K                     & $105.4^*$         \\ \hline
PCGAN  \cite{eiffert2020probabilistic}  & 66.2k                      & 18.63          \\ \hline
SR-LSTM \cite{zhang2019sr}    &  64.9K                       & 102.1          \\ \hline
CVM \cite{Scholler2019}    &  N/A                       & 29.45          \\ \hline
Social-STGCNN    & \textbf{7.6K}                       & $\textbf{4.2}^{*\dagger}$          \\ \hline
Ours-Probabilistic    & \textbf{7.6K}                       & $\textbf{4.2}^{*\dagger}$          \\ \hline
Ours-Deterministic   & \textbf{7.6K}                       & $\textbf{1.0}^\dagger$         \\ \hline
\end{tabular}
\caption{\textit{Comparison of model size and inference times for 8 timestep observation and 12 timestep prediction. Our proposed method has least number of parameters and fastest inference time of all models. * indicates 20 samples used. ${\dagger}$  indicates that graph encoding took place prior. Similar to \cite{mohamed2020social}, graph creation time is not included, which is 22.2 (ms) and 18.96 (ms) for ours and Social-STGCNN respectively. }}
\label{table_comp}
\end{table}

\begin{figure*}[tbp] 
\centering
\includegraphics[width=17.5cm,height=9cm]{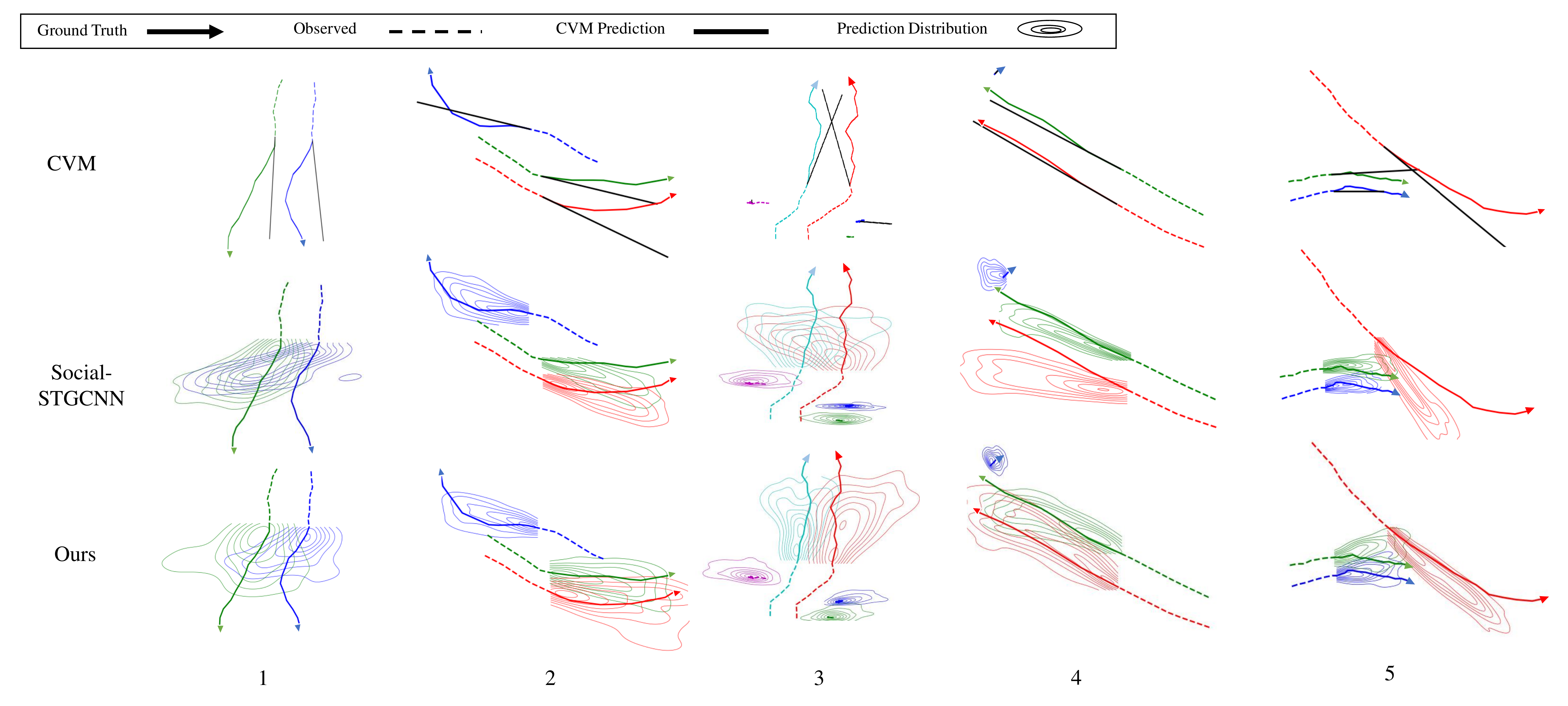}
\caption{\textit{Qualitative comparison with CVM and Social-STGCNN \cite{mohamed2020social} on interactive scenarios from ETH and UCY. Predicted distributions are shown as contour plots generated from 20 samples.  Scenario 1 shows two pedestrians walking in parallel. Scenario 2 and Scenarios 5 show a walking pedestrian approaching a group of pedestrians. Scenario 3 shows a group crossing several stationary pedestrians. Scenario 4 shows a group is approaching a stationary pedestrian.}}
\label{fig_quali}
\end{figure*}

\textbf{Deterministic Predictions:} \textit{Table. \ref{table_deterministic}} shows the results of testing the deterministic models on the ETH \cite{pellegrini2009you} and UCY \cite{lerner2007crowds} datasets, displaying both our deterministic model, and our probabilistic model that has been limited to outputting a single `most-likely' prediction. We see that directly training our model in a deterministic fashion as opposed to using the `most-likely' prediction of the probabilistic model results in increased prediction accuracy, an important consideration in time-critical tasks where only a single prediction without uncertainty is required.
Whilst SR-LSTM outperforms our deterministic model in both ADE and FDE, we see in \textit{Table. \ref{table_comp}} that SR-LSTM inference takes significantly longer, limiting the applicability for use in time-critical tasks in autonomous driving systems.
CVM also achieves good results, however as described in \cite{eiffert2020probabilistic} some subsets of the data used, such as ETH-Hotel, appear to involve significantly fewer pedestrian interactions suggesting that linear models should perform well. This appears to be the case when we focus on examples of pedestrian interaction from the datasets, as described in Section \ref{quali_eva} and shown in Fig. \ref{fig_quali}.

\textbf{Computation Speed Comparison:} We bench-mark inference time for 8 observed and 12 prediction timesteps running on an Nvidia GTX 1080Ti GPU and Intel Xeon E5v4 CPU in \textit{Table \ref{table_comp}}. Our proposed method and Social-STGCNN achieve the fastest computation speed among all models with the least number of parameters, suggesting that GCNN-based methods not only outperform state of the art probabilistic RNN-based methods but also requires much less computational resources. Furthermore, our approach also outperforms the CPU implemented Kalman filter based CVM method. 
As per \cite{mohamed2020social}, we do not include graph computation time due to the use of an non-optimized CPU implementation, however this process takes approximately 22ms, resulting in speeds still comparable to the next fastest PCGAN method when included. Whilst our proposed method has the same size and speed as Social-STGCNN \cite{mohamed2020social}, our method achieves superior performance on ADE and FDE metrics as \textit{Table \ref{table_pro_deter}} shows.

\subsection{Qualitative Evaluation} \label{quali_eva}
Comparisons in interactive scenarios, shown in \textit{Fig. \ref{fig_quali}}, confirms the robustness of our proposed method. We compare our proposed method with Social-STGCNN \cite{mohamed2020social} to demonstrate the benefit of our proposed NPA function during pedestrian encounters, and also to CVM \cite{Scholler2019} to highlight instances when linear predictions fail. 

\textbf{Parallel Walking} Generally, when a group of pedestrians are walking in parallel, they maintain similar walking speeds and direction. As it shows in Scenario 1 of \textit{Fig. \ref{fig_quali}} in which two pedestrians walk side by side, Social-STGCNN \cite{mohamed2020social} falsely predicts the pedestrians will collide with each other while our approach predicts similar distributions for each pedestrian. In Scenario 3 and Scenario 4, Social-STGCNN \cite{mohamed2020social} predicts the pedestrians will begin to diverge, whilst our method predicts future trajectory distributions with the same direction and speed. We can clearly see that CVM \cite{Scholler2019} fails in the situations when pedestrians change direction, or some noise exists, as in Scenario 3.
\medskip

\textbf{Collision Avoidance} Scenarios 4 and 5 show situations when pedestrians are approaching another stationary (4) or moving (5) pedestrian. Our proposed model predicts that both the approaching pedestrians and the stationary pedestrian will change heading slightly to avoid a collision, while Social-STGCNN falsely predicts greater incorrect deviation by approaching pedestrians. Similarly, when a single individual interacts with a group, it is difficult to predict their response, as illustrated in scenarios 2, 4, and 5. In (4), the stationary pedestrian is likely to move aside to avoid an approaching group. Our proposed method correctly predicts this motion while Social-STGCNN predicts an incorrect direction. Likewise, in (5), both CVM and Social-STGCNN \cite{mohamed2020social} predict a collision, while our predicted direction better follows the ground truth response.

\section{CONCLUSIONS}
In this paper, we demonstrate improved probabilistic pedestrian motion prediction using our proposed graph neural network with a novel graph attention function related to the proximity between pedestrians. We also show how this same model can be trained to improve deterministic performance when only a single prediction, and no measure of uncertainty is required. This use case is especially important to time-critical tasks in autonomous vehicle applications, which we explore through a comparison of model inference speeds, highlighting the advantage of our proposed approaches. Additionally, we address the need to use training datasets representative of real-world usage, proposing an automatically labelled dataset for autonomous vehicles to verify and evaluate pedestrian motion prediction models. Our work focuses on how pedestrians interact with each other to allow vehicles to navigate safely in crowds. In the future, we will extend our work to incorporate motion prediction for heterogeneous traffic agents allowing autonomous vehicles to safely navigate in more complex driving environments.






{\small
\bibliographystyle{unsrt}
\bibliography{egbib}

\begin{thebibliography}{10}

\bibitem{alahi2016social}
A.~Alahi, K.~Goel, V.~Ramanathan, A.~Robicquet, L.~Fei-Fei, and S.~Savarese.
\newblock Social {LSTM}: Human trajectory prediction in crowded spaces.
\newblock In {\em CVPR}, 2016.

\bibitem{zhang2019sr}
P.~Zhang, W.~Ouyang, P.~Zhang, J.~Xue, and N.~Zheng.
\newblock {SR-LSTM: State Refinement for LSTM towards Pedestrian Trajectory
  Prediction}.
\newblock In {\em CVPR}, 2019.

\bibitem{Vemula2018}
A.~Vemula, K.~Muelling, and J.~Oh.
\newblock {Social Attention : Modeling Attention in Human Crowds}.
\newblock In {\em ICRA}, 2018.

\bibitem{eiffert2020probabilistic}
S.~Eiffert, K.~Li, M.~Shan, S.~Worrall, S.~Sukkarieh, and E.~Nebot.
\newblock Probabilistic crowd gan: Multimodal pedestrian trajectory prediction
  using a graph vehicle-pedestrian attention network.
\newblock {\em IEEE Robotics and Automation Letters}, 5(4):5026--5033, 2020.

\bibitem{gupta2018social}
A.~Gupta, J.~Johnson, L.~Fei-Fei, S.~Savarese, and A.~Alahi.
\newblock Social {GAN}: Socially acceptable trajectories with generative
  adversarial networks.
\newblock In {\em CVPR}, 2018.

\bibitem{ivanovic2019}
B.~Ivanovic and M.~Pavone.
\newblock {The Trajectron: Probabilistic Multi-Agent Trajectory Modeling With
  Dynamic Spatiotemporal Graphs}.
\newblock In {\em ICCV}, 2019.

\bibitem{sophie2018}
A.~Sadeghian, V.~Kosaraju, A.~Sadeghian, N.~Hirose, S.~H. Rezatofighi, and
  S.~Savarese.
\newblock {SoPhie: An Attentive GAN for Predicting Paths Compliant to Social
  and Physical Constraints}.
\newblock In {\em CVPR}, 2018.

\bibitem{liang2019peeking}
J.~Liang, L.~Jiang, J.~C. Niebles, A.~G. Hauptmann, and L.~Fei-Fei.
\newblock Peeking into the future: Predicting future person activities and
  locations in videos.
\newblock In {\em CVPR}, 2019.

\bibitem{zhang2019stochastic}
L.~Zhang, Q.~She, and P.~Guo.
\newblock Stochastic trajectory prediction with social graph network.
\newblock {\em arXiv preprint arXiv:1907.10233}, 2019.

\bibitem{kosaraju2019social}
V.~Kosaraju, A.~Sadeghian, R.~Mart{\'\i}n-Mart{\'\i}n, I.~Reid, S.~H.
  Rezatofighi, and S.~Savarese.
\newblock Social-{BiGAT: Multimodal Trajectory Forecasting using Bicycle-GAN
  and Graph Attention Networks}.
\newblock In {\em NeurIPS}, 2019.

\bibitem{Rudenko2020}
A.~Rudenko, L.~Palmieri, M.~Herman, K.~M. Kitani, D.~M. Gavrila, and K.~O.
  Arras.
\newblock Human motion trajectory prediction: a survey.
\newblock {\em The International Journal of Robotics Research}, 39(8):895--935,
  2020.

\bibitem{Magdici2016}
S.~{Magdici} and M.~{Althoff}.
\newblock Fail-safe motion planning of autonomous vehicles.
\newblock In {\em ITSC}, 2016.

\bibitem{Salzmann2020}
T.~Salzmann, B.~Ivanovic, P.~Chakravarty, and M.~Pavone.
\newblock Trajectron++: Multi-agent generative trajectory forecasting with
  heterogeneous data for control.
\newblock {\em arXiv:2001.03093}, 2020.

\bibitem{mohamed2020social}
A.~Mohamed, K.~Qian, M.~Elhoseiny, and C.~Claudel.
\newblock Social-stgcnn: A social spatio-temporal graph convolutional neural
  network for human trajectory prediction.
\newblock In {\em CVPR}, pages 14424--14432, 2020.

\bibitem{Eiffert2020a}
S.~Eiffert, N.~Wallace, H.~Kong, N.~Pirmarzdashti, and S.~Sukkarieh.
\newblock {Path Planning in Dynamic Environments using Generative RNNs and
  Monte Carlo Tree Search}.
\newblock {\em ICRA}, 2020.

\bibitem{Fisac2018}
J.~F. Fisac, A.~Bajcsy, S.~L. Herbert, D.~Fridovich-Keil, S.~Wang, C.~Tomlin,
  and A.~D. Dragan.
\newblock {Probabilistically Safe Robot Planning with Confidence-Based Human
  Predictions}.
\newblock {\em RSS}, 2018.

\bibitem{velickovic2018graph}
P.~Veli{\v{c}}kovi{\'{c}}, G.~Cucurull, A.~Casanova, A.~Romero, P.~Li{\`{o}},
  and Y.~Bengio.
\newblock {Graph Attention Networks}.
\newblock {\em International Conference on Learning Representations}, 2018.

\bibitem{Eiffert2019}
S.~Eiffert and S.~Sukkarieh.
\newblock {Predicting Responses to a Robot's Future Motion Using Generative
  Recurrent Neural Networks}.
\newblock In {\em Australian Conference on Robotics and Automation (ACRA)},
  2019.

\bibitem{ma2019trafficpredict}
Y.~Ma, X.~Zhu, S.~Zhang, R.~Yang, W.~Wang, and D.~Manocha.
\newblock Trafficpredict: Trajectory prediction for heterogeneous
  traffic-agents.
\newblock In {\em AAAI}, 2019.

\bibitem{wang2020joint}
T.~Wang, L.~Liu, H.~Zhang, L.~Zhang, and X.~Chen.
\newblock Joint character-level convolutional and generative adversarial
  networks for text classification.
\newblock {\em Complexity}, 2020.

\bibitem{bai2018empirical}
S.~Bai, J.~Z. Kolter, and V.~Koltun.
\newblock An empirical evaluation of generic convolutional and recurrent
  networks for sequence modeling.
\newblock {\em arXiv preprint arXiv:1803.01271}, 2018.

\bibitem{vaswani2017attention}
A.~Vaswani, N.~Shazeer, N.~Parmar, J.~Uszkoreit, L.~Jones, A.~N. Gomez,
  {\L}.~Kaiser, and I.~Polosukhin.
\newblock Attention is all you need.
\newblock In {\em Advances in neural information processing systems}, 2017.

\bibitem{kipf2016semi}
T.~N. Kipf and M.~Welling.
\newblock Semi-supervised classification with graph convolutional networks.
\newblock {\em arXiv preprint arXiv:1609.02907}, 2016.

\bibitem{lerner2007crowds}
A.~Lerner, Y.~Chrysanthou, and D.~Lischinski.
\newblock Crowds by example.
\newblock In {\em Computer graphics forum}, volume~26, pages 655--664, 2007.

\bibitem{robicquet2016learning}
A.~Robicquet, A.~Sadeghian, A.~Alahi, and S.~Savarese.
\newblock Learning social etiquette: Human trajectory understanding in crowded
  scenes.
\newblock In {\em ECCV}, 2016.

\bibitem{yang2019top}
D.~Yang, L.~Li, K.~Redmill, and {\"U}.~{\"O}zg{\"u}ner.
\newblock Top-view trajectories: A pedestrian dataset of vehicle-crowd
  interaction from controlled experiments and crowded campus.
\newblock {\em Intelligent Vehicles Symposium}, 2019.

\bibitem{benfold2011stable}
B.~Benfold and I.~Reid.
\newblock Stable multi-target tracking in real-time surveillance video.
\newblock In {\em CVPR}, 2011.

\bibitem{zhou2012understanding}
B.~Zhou, X.~Wang, and X.~Tang.
\newblock Understanding collective crowd behaviors: Learning a mixture model of
  dynamic pedestrian-agents.
\newblock In {\em CVPR}, 2012.

\bibitem{sun2020scalability}
P.~Sun, H.~Kretzschmar, X~Dotiwalla, A.~Chouard, V.~Patnaik, P~Tsui~..., and
  D~Anguelov.
\newblock Scalability in perception for autonomous driving: Waymo open dataset.
\newblock In {\em CVPR}, 2020.

\bibitem{geiger2012we}
A.~Geiger, P.~Lenz, and R.~Urtasun.
\newblock {Are We Ready for Autonomous Driving? The KITTI Vision Benchmark
  Suite}.
\newblock In {\em CVPR}, 2012.

\bibitem{houston2020one}
J.~Houston, G.~Zuidhof, L.~Bergamini, Y.~Ye, A.~Jain, S.~Omari, V.~Iglovikov,
  and P.~Ondruska.
\newblock One thousand and one hours: Self-driving motion prediction dataset.
\newblock {\em arXiv preprint 2006.14480}, 2020.

\bibitem{he2016deep}
K.~He, X.~Zhang, S.~Ren, and J.~Sun.
\newblock Deep residual learning for image recognition.
\newblock In {\em {CVPR}}, 2016.

\bibitem{he2015delving}
K.~He, X.~Zhang, S.~Ren, and J.~Sun.
\newblock Delving deep into rectifiers: Surpassing human-level performance on
  imagenet classification.
\newblock In {\em ICCV}, 2015.

\bibitem{kingma2014adam}
D.~P. Kingma and J.~Ba.
\newblock Adam: A method for stochastic optimization.
\newblock {\em arXiv preprint arXiv:1412.6980}, 2014.

\bibitem{zhou2020developing}
W.~Zhou, J.S. Berrio~Perez, C.~De~Alvis, M.~Shan, S.~Worrall, J.~Ward, and
  E.~Nebot.
\newblock {Developing and Testing Robust Autonomy: The USyd Campus Dataset}.
\newblock {\em Intelligent Transportation Systems Magazine}, 2020.

\bibitem{kannala2004generic}
J.~Kannala and S.~Brandt.
\newblock A generic camera calibration method for fish-eye lenses.
\newblock In {\em ICPR}, 2004.

\bibitem{verma2019automatic}
S.~Verma, J.S. Berrio~Perez, S.~Worrall, and E.~Nebot.
\newblock {Automatic extrinsic calibration between a camera and a 3D Lidar
  using 3D point and plane correspondences}.
\newblock In {\em ITSC}, 2019.

\bibitem{gomez2020using}
F.~Gomez-Donoso, E.~Cruz, M.~Cazorla, S.~Worrall, and E.~Nebot.
\newblock {Using a 3D CNN for Rejecting False Positives on Pedestrian
  Detection}.
\newblock In {\em IEEE International Joint Conference on Neural Networks},
  2020.

\bibitem{2018yolov3}
J.~Redmon and A.~Farhadi.
\newblock Yolov3: An incremental improvement.
\newblock {\em arXiv preprint arXiv:1804.02767}, 2018.

\bibitem{garcia2016pointnet}
A.~Garcia-Garcia, F.~Gomez-Donoso, J.~Garcia-Rodriguez, S.~Orts-Escolano,
  M.~Cazorla, and J.~Azorin-Lopez.
\newblock Pointnet: A {3D} convolutional neural network for real-time object
  class recognition.
\newblock In {\em IEEE International Joint Conference on Neural Networks},
  2016.

\bibitem{Gomez_Donoso_2020_par3dnet}
F.~Gomez-Donoso, F.~Escalona, and M.~Cazorla.
\newblock Par3dnet: Using 3dcnns for object recognition on tridimensional
  partial views.
\newblock {\em Applied Sciences}, 10(10):3409, May 2020.

\bibitem{wojke2017simple}
N.~Wojke, A.~Bewley, and D.~Paulus.
\newblock Simple online and realtime tracking with a deep association metric.
\newblock In {\em {IEEE International Conference on Image Processing (ICIP)}},
  2017.

\bibitem{pellegrini2009you}
S.~Pellegrini, A.~Ess, K.~Schindler, and L.~Van~Gool.
\newblock You'll never walk alone: Modeling social behavior for multi-target
  tracking.
\newblock In {\em ICCV}, 2009.

\bibitem{Scholler2019}
C.~Sch{\"o}ller, V.~Aravantinos, F.~Lay, and A.~Knoll.
\newblock {What the Constant Velocity Model Can Teach Us About Pedestrian
  Motion Prediction}.
\newblock {\em ICRA}, 2020.

\end{thebibliography}
}

\end{document}